\pdfoutput=1

\documentclass[11pt]{article}

\usepackage[preprint]{acl}

\usepackage{times}
\usepackage{latexsym}
\usepackage{amsmath}
\usepackage[absolute,overlay]{textpos}

\usepackage{algorithm}
\usepackage{algorithmic}
\usepackage{rotating}

\usepackage{xcolor}

\definecolor{my_green}{RGB}{0, 176, 80}
\definecolor{my_red}{RGB}{255, 0, 0}
\usepackage{booktabs}
\usepackage{multirow} 

\usepackage[T1]{fontenc}

\usepackage[utf8]{inputenc}

\usepackage{microtype}

\usepackage{inconsolata}

\usepackage{graphicx}
\usepackage[most]{tcolorbox}

\usepackage{amsmath,amsfonts,bm}
\usepackage{xspace}

\def\eqref#1{equation~\ref{#1}}

\def\1{\bm{1}}

\mathchardef\mhyphen="2D

\DeclareMathAlphabet{\mathsfit}{\encodingdefault}{\sfdefault}{m}{sl}
\SetMathAlphabet{\mathsfit}{bold}{\encodingdefault}{\sfdefault}{bx}{n}

\newcommand{\method}{MEND}
\newcommand{\methodlong}{sy\textbf{M}metry-\textbf{EN}hance\textbf{D} (\textbf{MEND}) Data Augmentation}

\title{Your Language Model May Think Too Rigidly: Achieving Reasoning Consistency with Symmetry-Enhanced Training}

\author{
 \textbf{Yihang Yao\textsuperscript{1}},
 \textbf{Zhepeng Cen\textsuperscript{1}},
 \textbf{Miao Li\textsuperscript{1}},
 \textbf{William Han\textsuperscript{1}},
 \textbf{Yuyou Zhang\textsuperscript{1}},
 \textbf{Emerson Liu\textsuperscript{2}},
   \\
 \textbf{Zuxin Liu\textsuperscript{3}},
 \textbf{Chuang Gan\textsuperscript{4}},
 \textbf{Ding Zhao\textsuperscript{1}}
\\
 \textsuperscript{1}Carnegie Mellon University,
 \textsuperscript{2}AHN,
 \textsuperscript{3}Salesforce,
 \textsuperscript{4}UMass Amherst
\\
 {
   \{yihangya, zcen\}@andrew.cmu.edu
 }
}

\begin{document}
\maketitle
\begin{abstract}
Large Language Models (LLMs) have demonstrated strong reasoning capabilities across various tasks. However, even minor variations in query phrasing, despite preserving the underlying semantic meaning, can significantly affect their performance. To address this, we focus on enhancing LLMs' awareness of symmetry in query variations and propose \methodlong, a data-centric approach that improves the model's ability to extract useful information from context. Unlike existing methods that emphasize reasoning chain augmentation, our approach improves model robustness at the knowledge extraction stage through query augmentation, enabling more data-efficient training and stronger generalization to Out-of-Distribution (OOD) settings. Extensive experiments on both logical and arithmetic reasoning tasks show that \method\ enhances reasoning performance across diverse query variations, providing new insights into improving LLM robustness through structured dataset curation.
\end{abstract}

\section{Introduction}
Large Language Models (LLMs) have demonstrated superior performance across various reasoning tasks, including mathematical reasoning~\citep{YXLA2024-gsm1, qu2024recursive, lin2024rho, gou2023tora, shen2025satori}, code generation~\citep{chen2021evaluating, zhang2024diversity, zhang2023planning, dainese2024generating}, and autonomous system decision-making~\citep{yang2024agentoccam, sima2024drivelm}. Despite their ability to handle complex reasoning tasks, LLMs exhibit naive failure modes. For instance, they suffer from premise order sensitivity~\citep{chen2024premise, zhu2024dynamic}, the reverse-curse phenomenon~\citep{berglund2023reversal, golovneva2024reverse}, and distractibility~\citep{pmlr-v202-shi23a, zhu2024dyval}, making them vulnerable to variations in the natural language description of a query, even when the underlying semantic meaning remains unchanged.

\begin{figure*}[t]
    \centering
    \includegraphics[width=\linewidth]{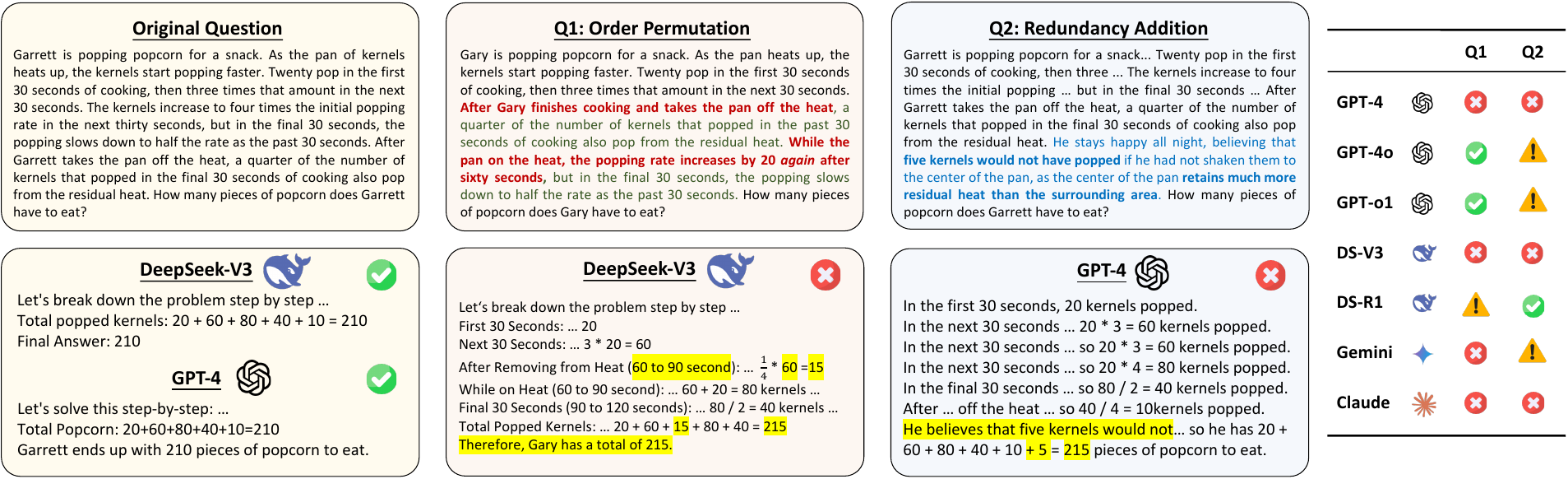}
    \caption{Failure examples of LLMs under surface form variations. Queries are modified from R-GSM~\citep{chen2024premise}. Table: The correctness for $10$ evaluations across different LLMs. \includegraphics[height=0.8em]{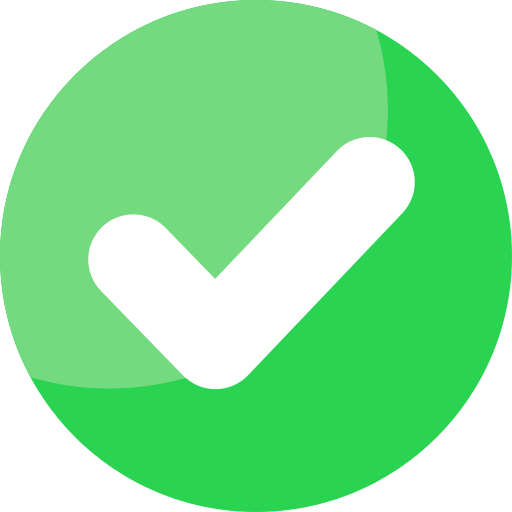}: all correct; \includegraphics[height=0.8em]{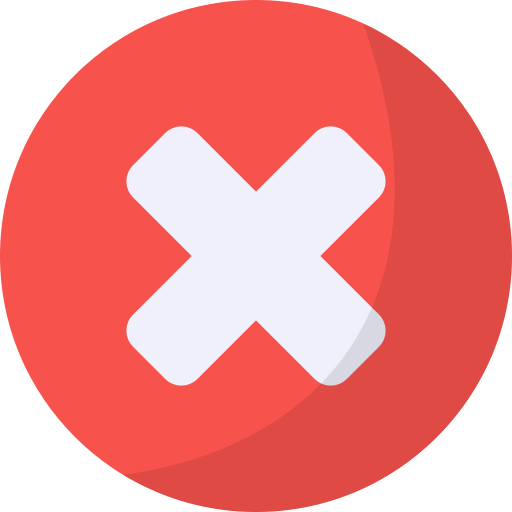}: all wrong; \includegraphics[height=0.8em]{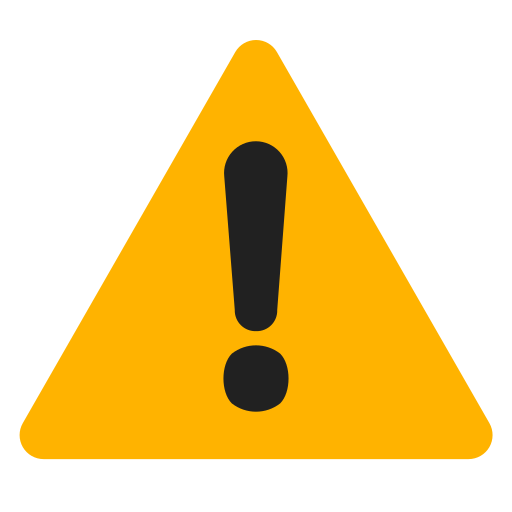}: error occurs. Full incorrect answers are provided in Appendix~\ref{subsection: close-source model evaluation}.}
    \label{fig: motivation}
    \vspace{-4mm}
\end{figure*}


A reasoning problem can be structurally decomposed into two layers~\citep{zhu2023dyval, zhou2024paraphrase}: the underlying layer represents the \textit{semantic meaning}, which includes the structure of reasoning chains and the knowledge required for logical deduction, while the upper layer consists of the \textit{surface form}, referring to the natural language description of the problem~\citep{zhou2024paraphrase}. The inability of LLMs to maintain consistent performance across different surface forms highlights the need for \textit{reasoning consistency}, which we define as the model's ability to generate consistent and correct answers from different query transformations of a problem with the same semantic meaning. In this work, we explore the question: \textit{how to improve LLMs' reasoning consistency across varying surface forms}.

Our key insight is that LLM reasoning problems maintain certain underlying structures, such as \textit{description symmetry}, which describes transformations of the query description that preserve its semantic meaning. Existing LLM post-training methods typically focus on improving the quality of training data for reasoning chains, such as using bootstrapping to enhance reasoning diversity~\citep{yu2023metamath}. However, these methods overlook the description symmetry in reasoning tasks, which naturally exists in surface form variations. As a result, they often suffer from overfitting and inconsistent reasoning with respect to different surface forms as shown in Figure~\ref{fig: motivation}.  
To address this issue, we propose \methodlong\ for LLM post-training. Our method augments the dataset with the symmetry information, enforcing the model's capability to better capture semantic meanings in different query variations, thereby improving post-training data efficiency and enhancing OOD generalization.
The main contributions of this work are:

\textbf{1. Formal analysis of the reasoning consistency problem.} We associate reasoning consistency in terms of an LLM’s ability to extract invariant knowledge, providing a formal framework for improving reasoning consistency.

\textbf{2. Introduction of \method\ to address the consistency of reasoning through post-training.} To our knowledge, this is the first systematic work that investigates post-training techniques to mitigate issues of consistency of reasoning.

\textbf{3. Extensive evaluation of \method\ in LLM reasoning tasks.} Our experimental results show that \method\ achieves superior sampling efficiency and generalizability in reasoning consistency with a significantly enhanced in-context knowledge extraction capability.

\section{Related Works}
\textbf{LLM Reasoning}: Recent studies highlight the remarkable reasoning capabilities of LLMs, exploring in-context learning~\citep{wei2022chain, yao2022react}, pre-training~\citep{YXLA2024-gsm1, YXLA2024-gsm2, shao2024deepseekmath, lightman2023let, lin2024not}, and post-training~\citep{ni2024exploring}. A key focus in training LLMs is the curation of high-quality datasets for instruction tuning~\citep{yue2023mammoth, liu2024augmenting, ni2024exploring}. Another approach involves leveraging LLM-generated datasets~\citep{wang2024math, cen2024bridging}, often with reinforcement learning~\citep{ouyang2022training, kumar2024training}. Both strategies highlight the essential role of high-quality data in enhancing LLM reasoning performance.
\\
\textbf{LLM Failure Modes:} Despite their advanced reasoning capabilities, LLMs exhibit surprising brittleness to variations in question descriptions with the same semantical meaning~\citep{chen2024premise}. For example, the \textit{Reversal Curse}~\citep{berglund2023reversal} refers to LLMs failing to generalize from statements like ``A is B'' to ``B is A.'' \textit{Premise Ordering} describes the performance degradation when the order of premises in a query differs from the order in their reasoning chains~\citep{chen2024premise}. \textit{Distractibility} is another failure mode in which LLMs' reasoning performance declines when irrelevant context is included in the query~\citep{shi2023large}. To mitigate these issues, researchers have explored inference-time scaling methods that paraphrase queries~\citep{zhou2024paraphrase} and post-training techniques~\citep{golovneva2024reverse} that reverses the order of tokens in training. However, a systematic analysis and solution to address these robustness challenges still remain an open problem.
\\
\textbf{Symmetry and Equivariant Learning:} Symmetry has been widely used to indicate rotationally
symmetric problems, such as image-input machine learning tasks~\citep{weiler2019general}. The definition of symmetry can also be extended to groups that preserve structured information while performing transformations~\citep{muglich2022equivariant}. Encoding data symmetries in the model training pipeline can improve both generalization and sample efficiency~\citep{wang2022mathrm}, an idea first proposed in G-Convolution~\citep{cohen2016group}. Recent works also incorporate equivariant learning in reinforcement learning~\citep{liu2023continual, wang2022mathrm} and robotics~\citep{yang2024equibot}. The symmetry in the training corpus of LLMs remains a widely unexplored area.

\section{Preliminary for LLM Reasoning}
\label{section: preliminary}
\textbf{DAG Representation.} Following previous works~\citep{zhu2023dyval, zhu2024dynamic, ye2024physics}, we formulate the reasoning task as a problem defined over a directed acyclic graph (DAG) representation: \( G = (V, E) \), where \( V = \{v_1, v_2, \dots, v_n\} \) represents the set of nodes, and \( E \subseteq V \times V \) represents the set of directed edges indicating dependencies or relationships between nodes. The root node \( v_r \in V \) corresponds to the target variable we aim to compute or reason about, leaf nodes \( v_l \in \mathcal{L} \subseteq V \) denote the variables with known values, and the other nodes \( v_i \in \mathcal{I} = V \setminus (\mathcal{L} \cup \{v_r\}) \) represent intermediate nodes that need to be computed by their parent nodes, which are denoted as \( \mathrm{Pa}(v_i) = \{v_j \mid (v_j, v_i) \in E\} \). Each directed edge \( (v_j, v_i) \in E \) indicates that the value of \( x_i \) depends on \( x_j \) and their quantitative relationship.
\\
\textbf{Ground Truth of Reasoning Chain:} The ground truth reasoning path for the reasoning chain is represented as a sequence of intermediate reasoning steps \( \{y_1, y_2, \dots, y_T, y\} \) to the final result \( y \), where \( y_t \) corresponds to the result of the \( t \)-th intermediate computation. This process is grounded in the structure of the directed acyclic graph \( G = (V, E) \). 
The reasoning process follows a topological sorting of \( G \), which is a linear ordering of its nodes such that for every directed edge \( (v_i, v_j) \in E \), node \( v_i \) appears before \( v_j \) in the ordering. 




\section{MEND Data Augmentation}

\label{section: method}
In this section, we first analyze the surface form and its transformations and reveal the symmetry in description queries. We then discuss LLMs' capability for reasoning consistency. Lastly, we introduce the data augmentation method to improve reasoning consistency and a probing tool for capability verification.

\subsection{Surface Form and Description Symmetry}
A reasoning problem consists of both a semantic meaning and a surface form~\citep{zhou2024paraphrase}. The semantic meaning is determined by the structure of the DAG and the values of its leaf nodes, capturing the core reasoning chain of the problem~\citep{pearl2009causality, velickovic2018geometric}. 
The surface form is the natural language description of the semantic meaning~\citep{zhou2024paraphrase}.
There exists a mapping \( f: \mathcal{Q} \to \mathcal{S} \) from the query space \( \mathcal{Q} \) to the semantic space \( \mathcal{S} \), which translates a natural language description into its underlying structured representation. This abstracts away linguistic or surface-level variations to focus on the task's core logical or mathematical structure~\cite{russell2021ai}. 

\begin{figure}[t]
    \centering
    \includegraphics[width=\linewidth]{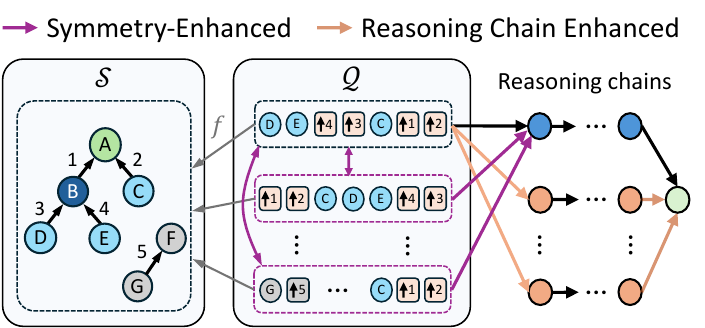}
    \caption{Overview of Symmetry-Enhanced Data Augmentation and its Comparison with Reasoning Chain Data Augmentation.}
    \label{fig: illustration}
    \vspace{-4mm}
\end{figure}

The symmetry relationship \( \sim \) on \( \mathcal{Q} \) is defined as:
\begin{equation}
    q_1 \sim q_2 \iff f(q_1) = f(q_2),
\end{equation}
where \( q_1, q_2 \in \mathcal{Q} \). That is, two query descriptions are symmetric if they correspond to the same semantics in \( \mathcal{S} \). 
We categorize symmetry relations in \( \mathcal{Q} \) into three types: \textbf{(1) Permutation}: reordering the provided information, \textbf{(2) Redundancy addition}: introducing irrelevant information as a distraction, and \textbf{(3) Surface-level variations}: paraphrasing questions at the linguistic level. Given the complexity of linguistic paraphrasing, this work primarily focuses on the first two types, leaving the third for future research.

\subsection{LLM Reasoning Consistency}
\label{subsection: LLM Reasoning Equivalence}
Given a query $q$, the reasoning process to find its ground-truth final answer $y$ of an auto-regressive LLM can be formulated as a conditioned generation process~\citep{xiang2025towards}:
\begin{equation}
\label{eq:math_format reasoning}
    \mathcal{P}(y | q) = \int_z \prod_{t=1}^T \mathcal{P}(y_t | y_{<t}, z, q) \mathcal{P}(z | q) dz,
\end{equation}
where \( \mathcal{P}(y | q) \) means the probability of generating right final answer and \( \mathcal{P}(z | q) \) represents \textbf{in-context knowledge extraction}, which extracts relevant knowledge \( z \) from \( q \). \( \mathcal{P}(y_t | y_{<t}, z, q) \) represents the conditioned generation to the final results step by step. 

We define the \textbf{reasoning consistency} as the ability of an LLM to maintain similar reasoning performance despite variations in the phrasing or description of a query $q$, as long as its underlying semantics remains unchanged, i.e.,
\begin{equation}
\label{eq: reasoning equivalence}
    \mathcal{P}(y \mid q_1) \approx \mathcal{P}(y \mid q_2), \quad q_1 \sim q_2
\end{equation}
Equation~\ref{eq:math_format reasoning} demonstrates that extracting high-quality representations $z$ from queries is crucial for ensuring reasoning consistency. In this work, we focus on \textit{invariant knowledge extraction}: improving the LLMs' capability to consistently extract useful $z$ despite the surface form changes.
\subsection{Symmetry-Enhanced Data Augmentation}
To improve representation extraction capability, we can adopt either \textit{model-centric} approaches, which focus on modifying model architectures, or \textit{data-centric} approaches, which aim to improve data quality. Given the complexity of modifying the architecture of pre-trained LLMs without compromising overall performance, we focus on a data-centric approach that encodes symmetry information in the post-training dataset and is compatible with most general LLM architectures.
We propose \methodlong as shown in Figure~\ref{fig: illustration}. Compared to previous methods that enhance the training chains in the training dataset~\citep{yu2023metamath}, we focus on query augmentation to impose a better understanding of symmetric structured information. Our method contains two parts:



\textbf{Step 1: Order Permutation.} We begin by splitting the original query $q$ into a list of partitions, using the newline character ("$\backslash$n") as the delimiter. Next, we shuffle the order of these partitions while preserving the semantic meaning, resulting in a reordered list of sentences. 

\textbf{Step 2: Redundancy Addition.} To add redundant information to the original question, we first randomly sample some new nodes and edges, which are not connected to the original DAG so that they will not contribute to the computation of the target node, and then generate the new premises accordingly following the previous template.
After this, we add the partition to a random position of the list, and then combine them together forming a query $q^\prime$ in the augmented dataset \( \mathcal{D}_\text{aug} \).

More details are presented in Alg.~\ref{algo:data_augmentation}. After the data augmentation, we finetune the language model using supervised fine-tuning (SFT) on the augmented dataset \( \mathcal{D}_\text{aug} \).

\begin{algorithm}[t]
\caption{\method\ }
{\bfseries Input:} \raggedright QA pair $(q, a)$ from the original dataset, augmentation times $K$ redundant information number $R$, separation delimiter SEP. \par
{\bfseries Output:} \raggedright Augmented dataset \( \mathcal{D}_\text{aug} \) for this QA pair \par
\begin{algorithmic}[1]
\STATE Initialize \( \mathcal{D}_\text{aug} \gets \{(q, a)\}\)
\FOR{\( i = 1, \dots, K \)}
\STATE \textcolor{blue}{\# Step 1: Order Permutation}
\STATE Divide the query into a segmentation list $L$ by SEP;
\STATE $\text{random.shuffle}(L)$
\STATE \textcolor{blue}{\# Step 2: Redundancy Addition}
    \FOR{\( j = 1, \dots, R \)}
        
        \STATE Randomly sample redundant nodes with random values and edges;
        \STATE Construct a redundant partition $l_r$ by applying template to the new nodes;
        \STATE $L$.append($l_r$);
    \ENDFOR
\STATE random.shuffle($L$);
\STATE $q^{\prime} \gets$ SEP.join($L$)
\STATE    \( \mathcal{D}_\text{aug} \gets \mathcal{D}_\text{aug} \cup \{(q^\prime, a)\} \)
\ENDFOR
\STATE {\bfseries Return:} \( \mathcal{D}_\text{aug} \)
\end{algorithmic} \label{algo:data_augmentation}
\end{algorithm}


\subsection{Knowledge Extraction Verification Tool}
\label{subsection: probing}
In the previous subsections, we proposed \method\ to enhance knowledge extraction capability and address reasoning consistency requirements.  
However, it remains unclear whether \method\ fundamentally improves reasoning capability or merely relies on memorization.  
In this section, we introduce the probing tool we use to evaluate the representation extraction capability, providing an explanation for the performance enhancement brought by \method.


Recent literature has revealed that in transformer-based LLMs, attention patterns are good indicators of whether LLMs can effectively retrieve useful information from queries~\citep{wang2024towards, hou2023towards}. Based on these findings, we utilize the attention of LLMs to analyze the knowledge extraction capability.

The core component of the attention-based probing is a binary classification task to determine whether a statement $q[i]$ in question query $q$ contains any useful information~\citep{hou2023towards}:
\begin{equation}
\label{eq:classification_probe}
    \mathcal{P}(q[i] | \boldsymbol{A}) \rightarrow \{0, 1\},
\end{equation}
where $\boldsymbol{A}=\{\boldsymbol{A}(l, h) \mid 1 \leq l \leq L ; 1 \leq h \leq H\}$ is the combination of attention weights across $L$ layers and $H$ heads after processing the question query $q$. The prediction result $1$ indicates $q[i]$ is useful in reasoning, and $0$ indicates $q[i]$ only contains irrelevant information. Following the previous work~\citep{hou2023towards}, we segment all tokens in a query into multiple groups and construct simplified attention $\boldsymbol{A}_\text{simp}$ by pooling in each group as the conditions to reduce the dimension of the attention weights:
\begin{equation}
\label{equ:simplified_attention}
    \mathcal{P}(q[i] | q, \boldsymbol{A}) \approx \mathcal{P}(q[i] | q, \boldsymbol{A}_\text{simp}),
\end{equation}
The previous attention-based probing method~\citep{hou2023towards} employs non-parametric algorithms, such as KNN, for query information retrieval. However, it struggles to accurately identify relevant premises, particularly in complex tasks or larger networks. This limitation arises because KNN treats all attention entries equally, overlooking the fact that the information at each position is inherently influenced by preceding content. As input length and network depth increase, this aggregation effect becomes more pronounced, further undermining the probing method's effectiveness. To address this issue, we adopt a linear probing approach based on logistic regression:
\begin{equation}
    \mathcal{P}(q[i] | \boldsymbol{A}_{\mathrm{simp}}) = \sigma(\boldsymbol{w}^\top \boldsymbol{A}_{\mathrm{simp}} + b),
\end{equation}
where $\boldsymbol{w}$ and $b$ are trainable variables. See more discussions on comparison between KNN-based and our probing methods in Appendix~\ref{subsection: appendix-model-probing}.


\section{Experiments}

In the experiments, we are going to answer the following research questions. (\textbf{RQ-1}) How do current LLMs perform in terms of reasoning consistency?
(\textbf{RQ-2}) How does \method\ benefit reasoning consistency?
(\textbf{RQ-3}) Why does \method\ improve reasoning consistency?
To answer these questions, we make the following experiment setup.

\subsection{Experiment setup}
\textbf{Tasks:} We evaluate our approach using two categories of tasks, logical reasoning and arithmetic reasoning, both selected from \texttt{PromptBench}~\citep{zhu2023dyval}. For simplicity, we focus on questions that can be represented as tree structures, where each non-leaf node has up to two parent nodes. The objective is to compute the value of the root node given the values of the leaf nodes and the computational rules from parent nodes to child nodes. Two example QA pairs are provided in Appendix~\ref{subsection:appendix-example-QA-pairs}.

In logical reasoning tasks, each node takes a boolean value, and the computational rules are sampled from $\{ \land (\texttt{AND}), \lor (\texttt{OR}), \neg(\texttt{NOT}) \}$. In arithmetic reasoning tasks, each node takes an integer value, with leaf node values sampled from $0\sim10$. The computational rules are drawn from $\{ +, -, \times, \square^2\}$. Given an underlying tree representation, \texttt{PromptBench} first assigns a random name to each node and then generates a natural language query using a fixed template, as shown in Appendix~\ref{subsection: example-qa-pairs}.

For the evaluation dataset, we create variations of surface forms with both order permutation and redundancy addition. For order permutation, we choose \texttt{Topological}: the topological sorting of the underlying DAG, which aligns with the reasoning chain~\citep{zhu2023dyval}, \texttt{Random}: random permutation of sentences in the question queries, and \texttt{Reversed}: the inversed order of \texttt{Topological}. We utilize the templates in the benchmark~\citep{zhu2023dyval} to ensure that the overall semantical meaning does not change with the order permutation. For redundancy addition, we add irrelevant descriptions in the query and set the number of redundant dependencies ranging from $0$ to $40$.

\begin{figure}[t ]
    \centering
    \includegraphics[width=1\linewidth]{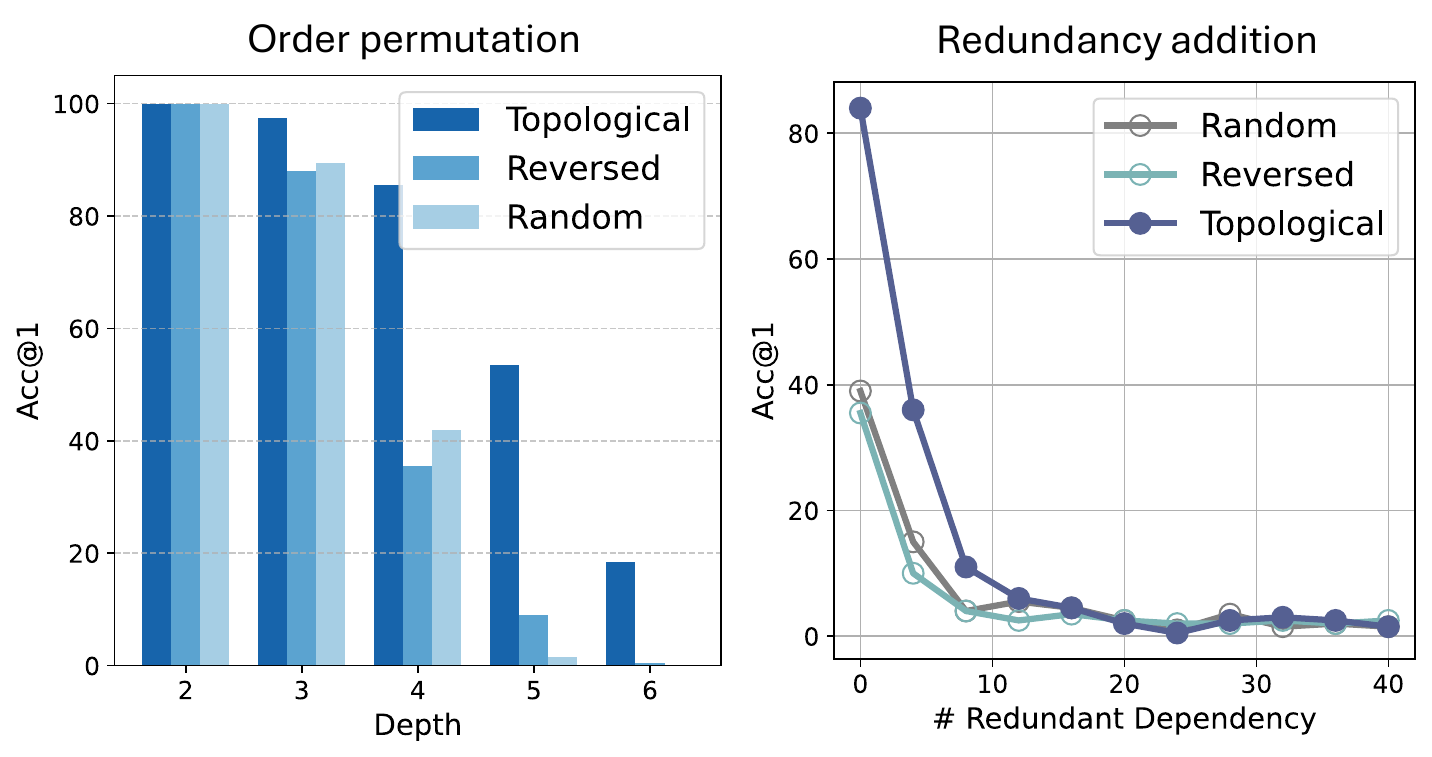}
    \caption{Accuracy evaluation of the DeepSeek-math-7B-base model on the arithmetic reasoning task with different surface forms. Left: results with different DAG depth; Right: results with different redundant information addition.}
    \label{fig: mathstral_evaluation}
    \vspace{-4mm}
\end{figure}

\textbf{Models:} For all experiments, we utilize the Llama-3.2 1B and 3B models~\citep{dubey2024llama3} as our base models unless specified otherwise.

\textbf{Baselines:} We compare our method with three types of baselines: \textbf{(1) Post-training:} \texttt{Vanilla}: trains base models on the vanilla dataset using SFT; and \texttt{RC-Aug}: utilizes the reasoning chain (RC) augmented dataset for SFT~\citep{yu2023metamath}. Specifically, we augment the answers with different topological orderings while keeping the query unchanged; \textbf{(2) Inference scaling:} \texttt{SCoP}-$k$~\citep{zhou2024paraphrase}, first paraphrases the question prompts $k$ times, then performs reasoning based on these paraphrases, and obtains the final answer via majority voting. It is combined with \texttt{vanilla} models unless specified otherwise. \textbf{(3) Ablation:} In addition to these baselines, we also create a variant of our method called \texttt{MEND-RC} that uses \texttt{\method} to transform the queries, while augmenting the dataset with more reasoning chains. We evaluate the models with greedy generation (temperature=$0$) and report the averaged accuracy on $200$ testing samples on every evaluation dataset unless specified otherwise.



\subsection{Reasoning-Consistency Evaluation}
\label{subsection: Reasoning-Equivalence Evaluation}
To answer \textbf{RQ1}, we evaluate several LLMs with query variations and present the reasoning accuracy in Figure~\ref{fig: mathstral_evaluation}. Due to space limitations, we defer the additional results to Appendix~\ref{subsection:appendix-open-source models}. We observe two key findings: (1) The topological order improves the accuracy of generation, while adversarial permutations, including random and reversed orders, degrade performance. Furthermore, the harder the reasoning task (i.e., greater DAG depth), the more pronounced the performance drop, consistent with previous findings for closed-source LLMs~\citep{chen2024premise}.
(2) {Redundant information in the question query negatively affects reasoning performance.} As the amount of irrelevant information increases, the model's reasoning accuracy decreases.

Both observations indicate that open-source models still suffer from \textbf{overfitting to topological order and redundancy-free queries}, leading to poor performance in reasoning consistency tasks. This issue further motivates the development of techniques to enhance reasoning-consistency robustness in LLMs.


\begin{figure*}[ht]
    \centering
    \includegraphics[width=\linewidth]{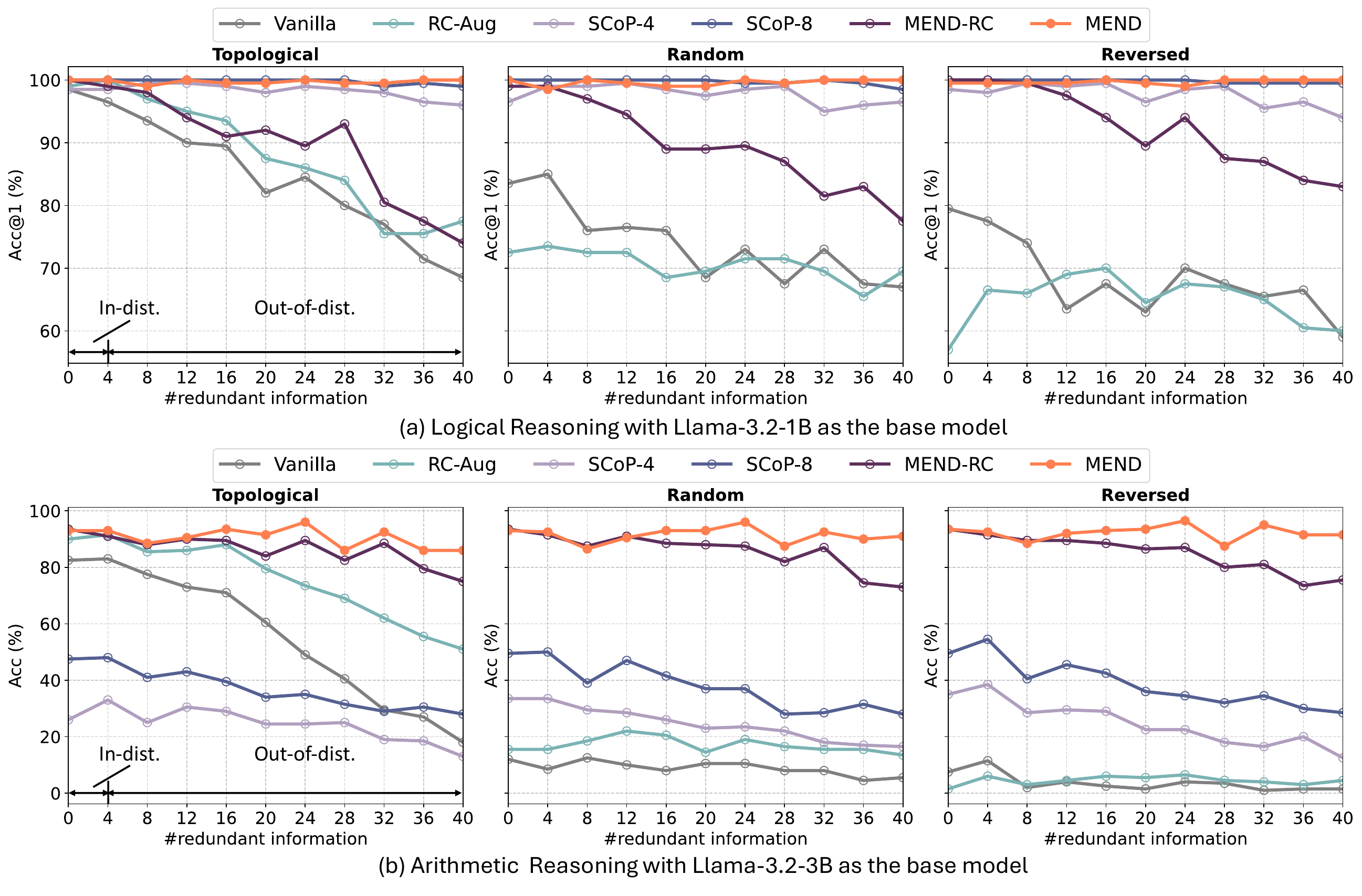}
    \caption{Evaluations with respect to different query variations. Each figure refers to one permutation order type, the x-axis represents the number of redundancies of the test set, and the y-axis represents the accuracy of final answers. For each dataset, we report the accuracy value over a dataset with a size of $200$. 
    }
    \label{fig: Math reasoning}
    \vspace{-2mm}
\end{figure*}

\subsection{Reasoning-Consistency Enhancement}
\label{subsection:exp reasoning-equivalence enhancement}
To answer \textbf{RQ2}, we present the main experiments with two main parts: reasoning consistency on order permutation and redundancy addition.

\textbf{Order permutation}: We train the LLM on arithmetic questions with difficulty level $2$ and $3$ while testing its performance on difficulty level $1\sim4$ with three different permutation orders: \texttt{Topological}, \texttt{Random} and \texttt{Reversed}. The evaluation results on Llama-3.2-3B are listed in Table~\ref{tab: permutation-topological}. More results with different base models are available in the Appendix~\ref{subsection:appendix-permutation-order-experiments}.

\begin{table}[htbp]
  \centering
  \caption{Accuracy (\%) evaluation on datasets with different permutation order and base model as Llama-3.2-3B. The difficulty level is defined by the number of reasoning steps for ground-truth reasoning chains. The number in the parentheses indicates the performance comparison with the \texttt{Vanilla} method. 
  \textbf{Bold}: the method with the best performance.}
  \resizebox{1.\linewidth}{!}{
    \begin{tabular}{ccccccc}
    \toprule
    \multirow{2}[4]{*}{Order} & \multirow{2}[4]{*}{Method} & \multicolumn{4}{c}{Difficulty Level} & \multirow{2}[4]{*}{Avg.} \\
\cmidrule{3-6}          &       & 1     & 2     & 3     & 4     &  \\
    \midrule
    \multirow{6}[2]{*}{\begin{sideways}Topological\end{sideways}} & Vanilla & 100.0  & 99.0  & 79.5  & 32.0  & 77.6  \\
          & RC-Aug & 100.0  & 98.5  & 87.0  & 33.0  & 79.6 (\textcolor{my_green}{+2.0}) \\
          & SCoP-2 & 100.0  & 82.5  & 16.0  & 1.5   & 50.0 (\textcolor{my_red}{-27.6}) \\
          & SCoP-4 & 100.0  & 94.5  & 28.5  & 3.5   & 56.6 (\textcolor{my_red}{-21.0}) \\
          & SCoP-8 & 100.0  & 100.0  & 49.5  & 9.0   & 64.6 (\textcolor{my_red}{-13.0}) \\
          & MEND  & 97.5  & 99.5  & 88.5  & 37.5  & \textbf{80.8 (\textcolor{my_green}{+3.2})} \\
    \midrule
    \multirow{6}[2]{*}{\begin{sideways}Random\end{sideways}} & Vanilla & 99.0  & 65.5  & 10.5  & 0.0   & 43.8  \\
          & RC-Aug & 100.0  & 62.5  & 13.0  & 1.0   & 44.1 (\textcolor{my_green}{+0.3}) \\
          & SCoP-2 & 99.0  & 77.5  & 15.0  & 3.0   & 48.6 (\textcolor{my_green}{+4.8}) \\
          & SCoP-4 & 100.0  & 96.5  & 26.0  & 4.5   & 56.8 (\textcolor{my_green}{+13.0}) \\
          & SCoP-8 & 100.0  & 99.5  & 38.5  & 12.5  & 62.6 (\textcolor{my_green}{+18.8}) \\
          & MEND  & 97.5  & 99.5  & 85.5  & 37.5  & \textbf{80.0 (\textcolor{my_green}{+36.2})} \\
    \midrule
    \multirow{6}[2]{*}{\begin{sideways}Reversed\end{sideways}} & Vanilla & 99.5  & 45.0  & 5.5   & 1.0   & 37.8  \\
          & RC-Aug & 100.0  & 30.5  & 0.5   & 0.5   & 32.9 (\textcolor{my_red}{-4.9}) \\
          & SCoP-2 & 99.5  & 85.0  & 20.5  & 2.5   & 51.9 (\textcolor{my_green}{+14.1}) \\
          & SCoP-4 & 100.0  & 97.0  & 30.0  & 4.5   & 57.9 (\textcolor{my_green}{+20.1}) \\
          & SCoP-8 & 100.0  & 100.0  & 42.5  & 8.0   & 62.6 (\textcolor{my_green}{+24.8}) \\
          & MEND  & 97.5  & 98.5  & 86.0  & 41.0  & \textbf{80.8 (\textcolor{my_green}{+43.0})} \\
    \bottomrule
    \end{tabular}%
    }
  \label{tab: permutation-topological}%
\end{table}%

Table~\ref{tab: permutation-topological} shows that training solely on the topological order corpus results in overfitting - it achieves superior performance on the \texttt{Topological} test set but suffers significant degradation on the \texttt{Random} and \texttt{Reversed} sets. \texttt{RC-Aug} provides only minor improvements in overall performance. \texttt{SCoP-k}, the inference-time scaling baseline, improves performances on the \texttt{Reversed} and \texttt{Random} sets but degrades on the \texttt{Topological} set. In contrast, our method demonstrates consistently strong performances on all evaluation datasets, effectively mitigating overfitting issues.

\textbf{Redundancy Addition}: To evaluate the robustness of LLMs against redundancy, we add $0$ to $4$ redundant premises to each query in the SFT dataset and test the LLMs' performance on questions with up to 40 redundant premises, with useful information provided in three permutation orders. The evaluation results are presented in Figure~\ref{fig: Math reasoning}. Additional results with different model sizes are provided in Appendix~\ref{subsection:supplementary-exp-redundancy}.

Figure~\ref{fig: Math reasoning} shows the \texttt{Vanilla} method performs well only on in-distribution samples with \texttt{Topological} order. As redundancy increases in the OOD setting, prediction accuracy declines sharply, indicating overfitting. When SFT with the \texttt{RC-Aug} dataset, the model shows noticeable performance improvements under in-distribution conditions, likely due to an enhanced capability for reasoning chain generation, as it learns from more reasoning chain data during post-training. Despite this improvement, \texttt{RC-Aug} still struggles to generalize well as the level of redundancy increases.  
The inference-time scaling method, \texttt{SCoP}-$k$, results in more consistent performance across surface variations. We observe its significant improvements in OOD settings, including \texttt{Random} and \texttt{Reversed} permutation orders and increased redundancy. However, the improvement is still limited, especially for more challenging arithmetic reasoning task. Additionally, it even shows performance degradation on the \texttt{Topological} dataset, where information is originally provided in order and disturbed in paraphrases.

Results from Figure~\ref{fig: Math reasoning} also shows \method, consistently outperforms all baselines across different permutation orders and levels of redundancy. These results demonstrate that diversifying question queries for SFT sufficiently enhances the ability of models to capture symmetry information across various surface forms. Consequently, our method not only performs well on in-distribution evaluation samples but also shows strong generalizability to OOD cases with unseen redundancy.  
We also conduct an ablation study with \texttt{MEND-RC}, where we observe a performance degradation when replacing the data augmentation direction from queries to reasoning chains, highlighting the necessity of enhancing query symmetry information for SFT.


\begin{figure}[t]
    \centering
    \includegraphics[width=\linewidth]{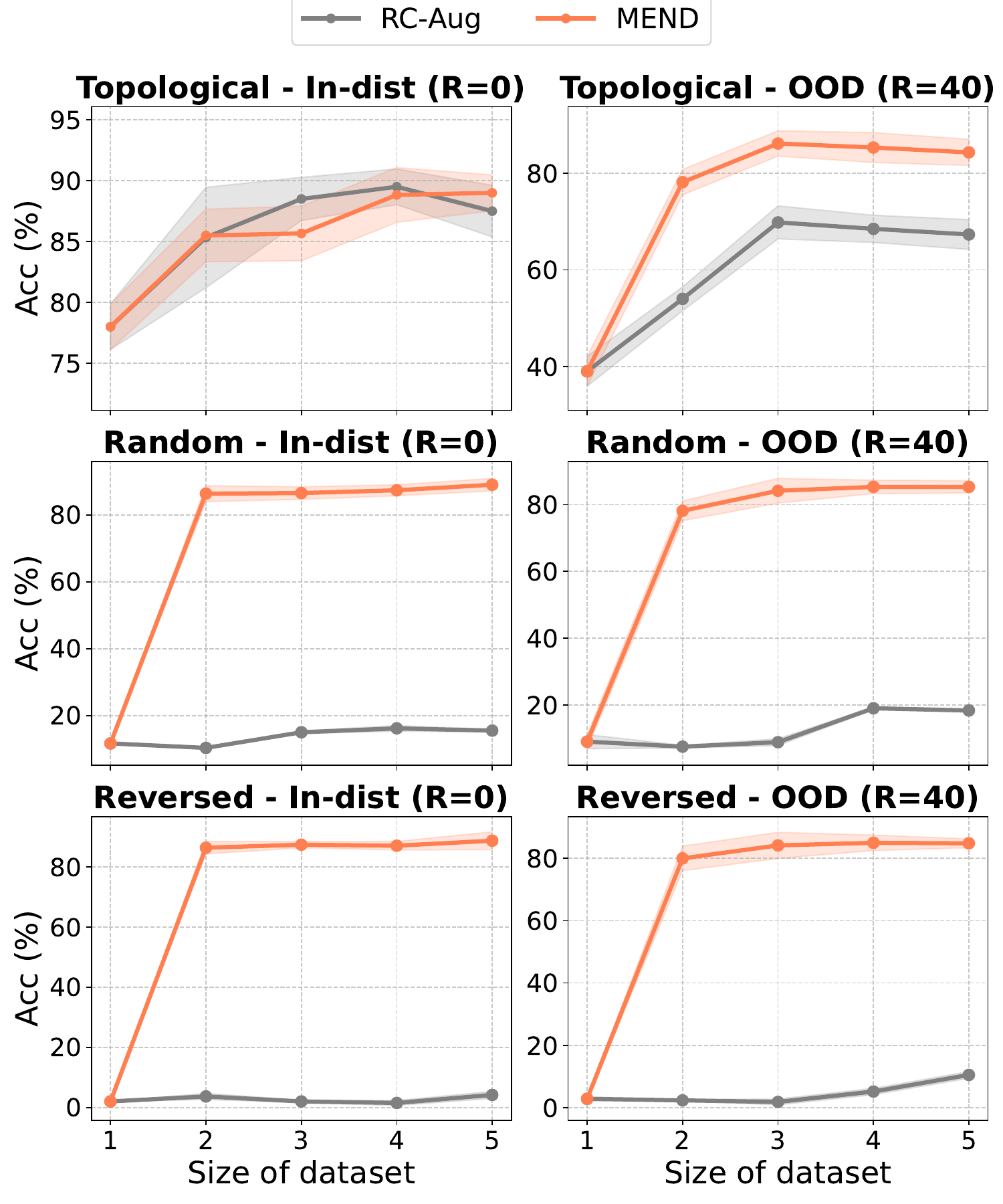}
    \caption{Data efficiency evaluation. R in figure titles indicates the number of redundancy in the query. The size of dataset $=1$ indicates using the original dataset for SFT. All plots are averaged among $3$
random seeds with temperature$=1$. The solid line is the mean value, and the light shade represents the first standard deviation.}
    \label{fig: data efficiency}
    \vspace{-2mm}
\end{figure}

\begin{table}[htbp]
  \centering
  \caption{VoV evaluation for arithmetic tasks with {Llama-3.2-3B} model. $(\downarrow)$: The lower, the better for reasoning equivalence. We normalize the scale with respect to the variance of \texttt{Vanilla} method. The VoV values for \texttt{Vanilla} are $195.55$, and $966.22$.}
  \resizebox{1.\linewidth}{!}{
    \begin{tabular}{cccccc}
    \toprule
     & Vanilla & RC-Aug & SCoP-8 & MEND-RC & MEND \\
    \midrule
    $\mathrm{VoV}_\mathbf{o}$ $(\downarrow)$ & 1.00 & 3.70E+00 & 3.28E-01 & 2.03E-01 & 4.47E-02 \\
    $\mathrm{VoV}_\mathbf{r}$ $(\downarrow)$ & 1.00 & 1.55E+00 & 3.81E-03 & 3.53E-03 & 2.23E-03 \\
    \bottomrule
    \vspace{-6mm}
    \end{tabular}%
    }
  \label{tab:VoV}%
\end{table}%


\textbf{Reasoning consistency:} We adopt the Variance of Variations (VoV)~\citep{zhou2024paraphrase} to quantitatively evaluate the reasoning consistency:  
\begin{equation}
\mathrm{VoV}_{\mathbf{f}}=\operatorname{Var}_{\mathbf{f}}(\mathrm{acc}(p)),\ {\mathbf{f}} \in \{\mathbf{o}, \mathbf{r}\},
\end{equation}
where $\mathrm{VoV}_\mathbf{o}$ and $\mathrm{VoV}_\mathbf{r}$ are the variance of prediction accuracy with respect to different permutation orders and different levels of redundancy. Smaller $\mathrm{VoV}$ values indicate better reasoning consistency. 

Table~\ref{tab:VoV} shows the VoV values for the arithmetic reasoning task in Figure~\ref{fig: Math reasoning}. The \texttt{Vanilla} and \texttt{RC-Aug} method show very large VoV values, indicating poor reasoning consistency and poor capability to capture the symmetry information. The \texttt{SCoP} method successfully reduces the variance since the final results are obtained through majority voting from diverse query paraphrases. Our method also demonstrates strong performance in reducing variance, indicating that we effectively mitigate reasoning consistency issues during post-training without sacrificing inference-time efficiency. 

\textbf{Data Efficiency:} We use different amounts of data for SFT and present the results of the arithmetic reasoning task with {Llama-3.2-1B} model in Figure~\ref{fig: data efficiency}. For the \texttt{RC-Aug} baseline, we observe that increasing the size of the augmented dataset leads to a rapid accuracy improvement in in-distribution conditions but a much slower improvement in OOD settings. This indicates that while \texttt{RC-Aug} can improve performance, it is not efficient in OOD scenarios. 
In contrast, \method\ consistently outperforms the baselines across different amounts of SFT data in OOD settings, demonstrating superior data efficiency and generalizability.

\begin{figure}[t]
    \centering
    \includegraphics[width=0.85\linewidth]{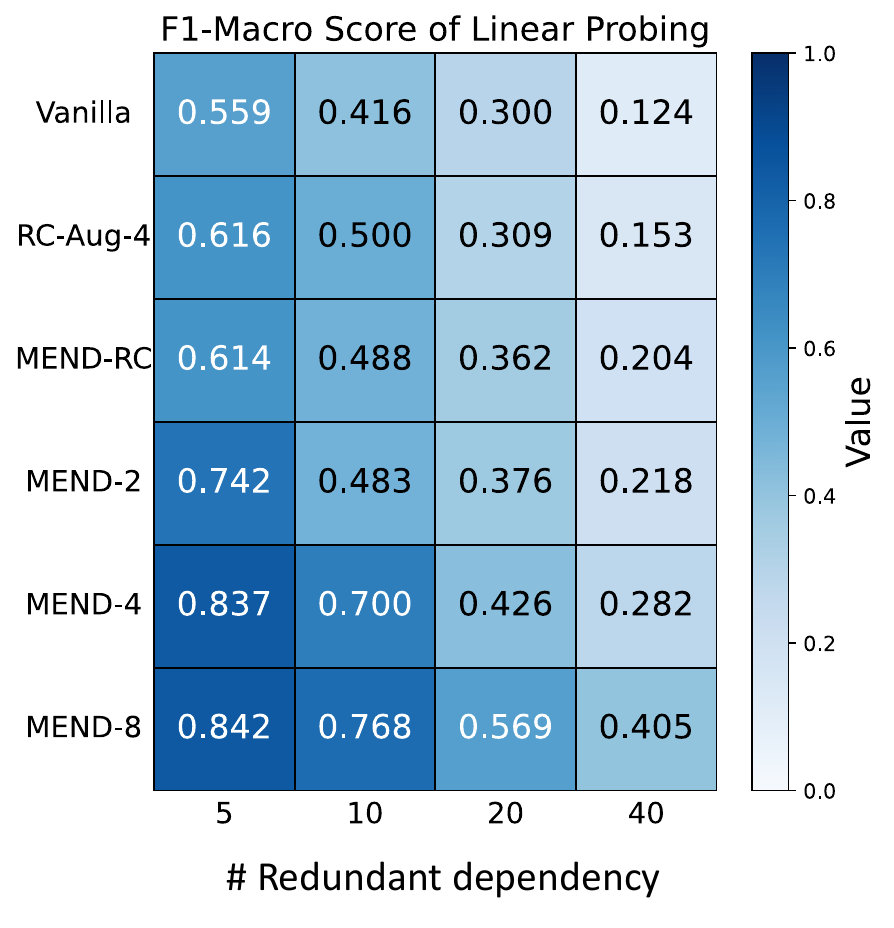}
    \vspace{-3mm}
    \caption{F1-macro score of linear probing on the logical reasoning task with the base model as \texttt{Llama-3.2-1B}. \texttt{Method}-$n$ indicates the dataset size is $n$ times the original one after augmentation.}
    \label{fig: Linear probing}
    \vspace{-4mm}
\end{figure}

\subsection{Reasoning-Consistency Verification} 
In this subsection, we aim to answer \textbf{RQ3} by verifying whether \method\ captures the structured information in question queries. We utilize the probing method described in Section~\ref{subsection: probing} and report the F1-Macro scores of prediction (\ref{eq:classification_probe}) as the evaluation metric~\citep{hou2023towards}. A higher score means a better ability to retrieve relevant information from the input queries.

The results are presented in Figure~\ref{fig: Linear probing}. We observe that the baseline methods \texttt{RC-Aug} and \texttt{MEND-RC} do not show significant improvement in detecting useful information compared to the \texttt{Vanilla} model.  
In contrast, the proposed method \method\ achieves a much higher score than the baselines. This confirms that by enhancing query symmetry, \method\ captures more structured information and significantly improves LLMs' \textit{in-context knowledge extraction} capability for reasoning tasks. The probing results also explain the OOD generalizability improvement of \method, as it enhances the LLMs' ability to understand OOD queries. A comparison between our linear probing and KNN probing is provided in Appendix~\ref{subsection: appendix-model-probing}.

\section{Conclusion}
In this work, we addressed the problem of reasoning equivalence in LLMs with respect to diverse surface form variations corresponding to the same semantic meaning. We proposed \methodlong, enhancing reasoning equivalence by applying structured query-level transformations. Our method improves reasoning robustness at the knowledge extraction stage by enforcing the query symmetry in the SFT dataset.
Experiments demonstrate that \method\ achieves superior sampling efficiency and generalizability in reasoning tasks. Although one potential risk is that the misuse of this work in real-world scenarios can cause unexpected damage, we hope our findings can provide a foundation for future research on improving LLM reasoning consistency through structured dataset curation. 

\section*{Limitations}
While \method\ demonstrates improved performance of reasoning consistency under query variations, we primarily evaluate it on arithmetic and logical reasoning tasks. As a result, its potential and effectiveness on more complex tasks or domains such as specialized scientific reasoning and code generation remain underexplored. 



\bibliography{custom}

\clearpage
\appendix

\section{Supplementary Experiments}
\label{sec:appendix-exp}
In this section, we provide some supplementary experiments that are omitted in the main context due to the space limit.

\begin{figure}[h]
    \centering
    \includegraphics[width=1\linewidth]{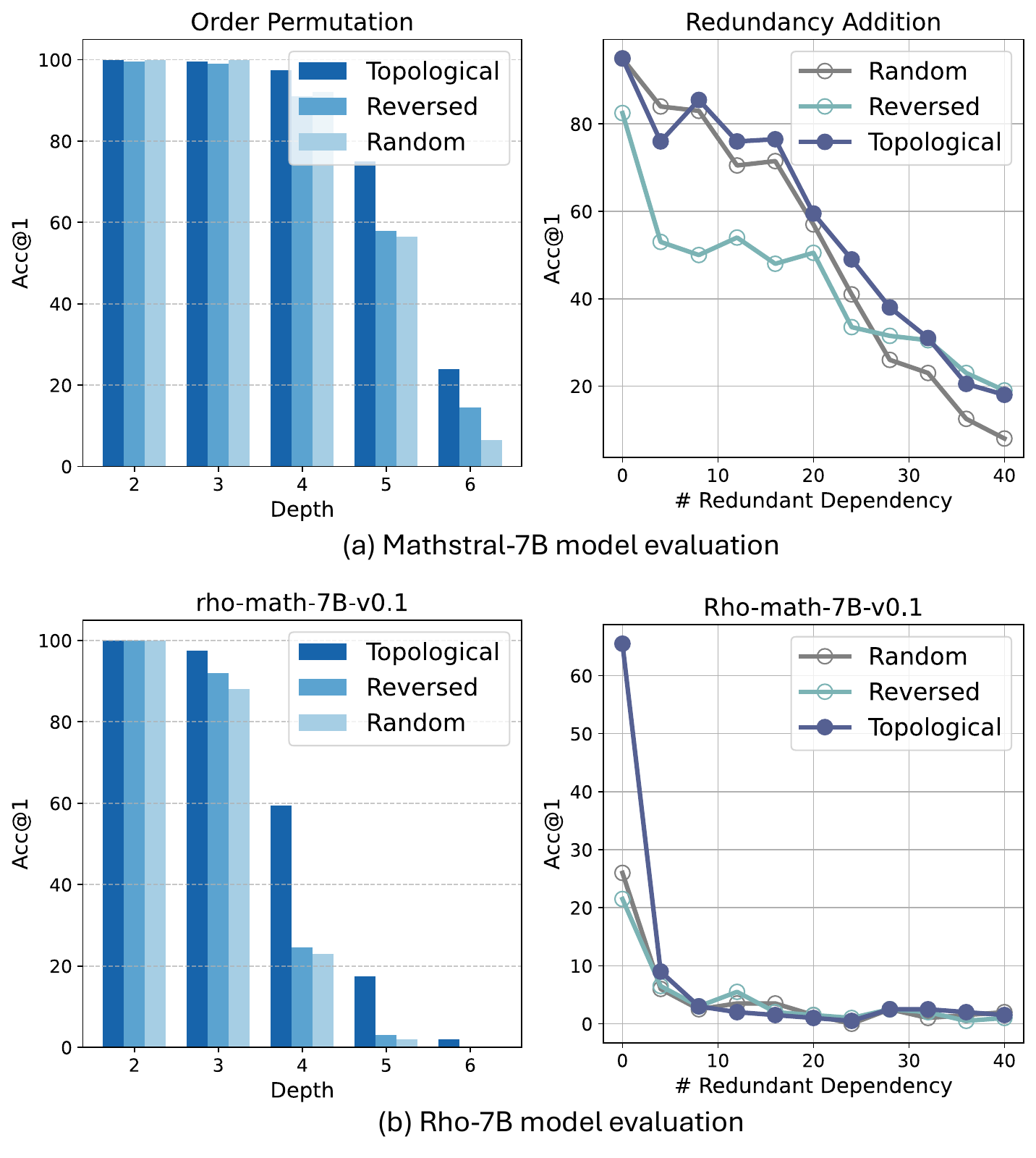}
    \caption{Evaluation of the {mathstral-7B} and {rho-math-7b-v0.1} model.}
    \label{fig: open model evaluation}
\end{figure}

\subsection{Supplementary evaluation of open-source models}
\label{subsection:appendix-open-source models}
In addition to the evaluation of DeepSeek model we presented in section~\ref{subsection: Reasoning-Equivalence Evaluation}, we also provide the evaluation results of {Mathstral-7B} and \texttt{Rho-math-7b-v0.1}. The results of the $8$-shot accuracy evaluation are presented in Figure~\ref{fig: open model evaluation}. The findings are consistent with those in section~\ref{subsection: Reasoning-Equivalence Evaluation} that LLMs are vulnerable to order permutation and redundancy addition.

\subsection{Supplementary results about permutation order experiments}
\label{subsection:appendix-permutation-order-experiments}
\begin{table}[t]
  \centering
  \caption{Accuracy (\%) evaluation on datasets with different permutation order and base model as Llama-3.2-1B. The difficulty level is defined by the number of reasoning steps for ground-truth reasoning chains. The number in the parentheses indicates the performance comparison with the vanilla method. \textcolor{my_green}{Green}: performance improvement; \textcolor{my_red}{Red}: performance degradation. \textbf{Bold}: the method with best performance.}
    \resizebox{1.\linewidth}{!}{\begin{tabular}{ccccccc}
    \toprule
    \multirow{2}[4]{*}{Order} & \multirow{2}[4]{*}{Method} & \multicolumn{4}{c}{Difficulty Level} & \multirow{2}[4]{*}{Avg.} \\
\cmidrule{3-6}          &       & 1     & 2     & 3     & 4     &  \\
    \midrule
    \multirow{6}[2]{*}{\begin{sideways}Topological\end{sideways}} & Vanilla & 99.0  & 99.0  & 75.5  & 23.0  & 74.1  \\
          & RC-Aug & 100.0  & 100.0  & 84.5  & 24.5  & \textbf{77.3} \textcolor{my_green}{\textbf{(+3.2)}} \\
          & SCoP-2 & 100.0  & 82.5  & 19.0  & 2.0   & 50.9 \textcolor{my_red}{(-23.2)} \\
          & SCoP-4 & 100.0  & 92.5  & 29.0  & 4.5   & 56.5 \textcolor{my_red}{(-17.6)} \\
          & SCoP-8 & 100.0  & 99.0  & 40.5  & 9.0   & 62.1 \textcolor{my_red}{(-12.0)} \\
          & MEND  & 100.0  & 100.0  & 81.0  & 11.5  & 73.1 \textcolor{my_red}{(-1.0)} \\
    \midrule
    \multirow{6}[2]{*}{\begin{sideways}Random\end{sideways}} & Vanilla & 99.5  & 57.0  & 12.0  & 1.0   & 42.4  \\
          & RC-Aug & 100.0  & 65.0  & 12.0  & 0.5   & 44.4 \textcolor{my_green}{(+2.0)} \\
          & SCoP-2 & 100.0  & 76.5  & 14.0  & 2.5   & 48.3 \textcolor{my_green}{(+5.9)} \\
          & SCoP-4 & 100.0  & 94.0  & 27.5  & 4.0   & 56.4 \textcolor{my_green}{(+14.0)} \\
          & SCoP-8 & 100.0  & 98.5  & 38.0  & 7.5   & 61.0 \textcolor{my_green}{(+18.6)} \\
          & MEND  & 100.0  & 100.0  & 80.5  & 7.5   & \textbf{72.0} \textcolor{my_green}{\textbf{(+29.6)}} \\
    \midrule
    \multirow{6}[2]{*}{\begin{sideways}Reversed\end{sideways}} & Vanilla & 98.5  & 8.5   & 0.0   & 0.5   & 26.9  \\
          & RC-Aug & 100.0  & 29.0  & 0.5   & 0.0   & 32.4 \textcolor{my_green}{(+5.5)} \\
          & SCoP-2 & 100.0  & 79.0  & 22.5  & 3.5   & 51.3 \textcolor{my_green}{(+24.4)} \\
          & SCoP-4 & 100.0  & 95.5  & 30.0  & 4.5   & 57.5 \textcolor{my_green}{(+30.6)} \\
          & SCoP-8 & 100.0  & 98.5  & 41.0  & 3.0   & 60.6 \textcolor{my_green}{(+33.7)} \\
          & MEND  & 100.0  & 100.0  & 81.0  & 4.5   & \textbf{71.4} \textcolor{my_green}{\textbf{(+44.5)}} \\
    \bottomrule
    \end{tabular}%
    }
  \label{tab:appendix-permutation}%
\end{table}%
The supplementary experiment results about arithmetic reasoning tasks using {Llama-3.2-1B} model are shown in Table~\ref{tab:appendix-permutation}. The observations are basically the same as those in Section~\ref{subsection:exp reasoning-equivalence enhancement}: the proposed \method\ achieves the best performance in OOD scenarios against all the baseline methods.


\begin{figure*}[t]
    \centering
    \includegraphics[width=1\linewidth]{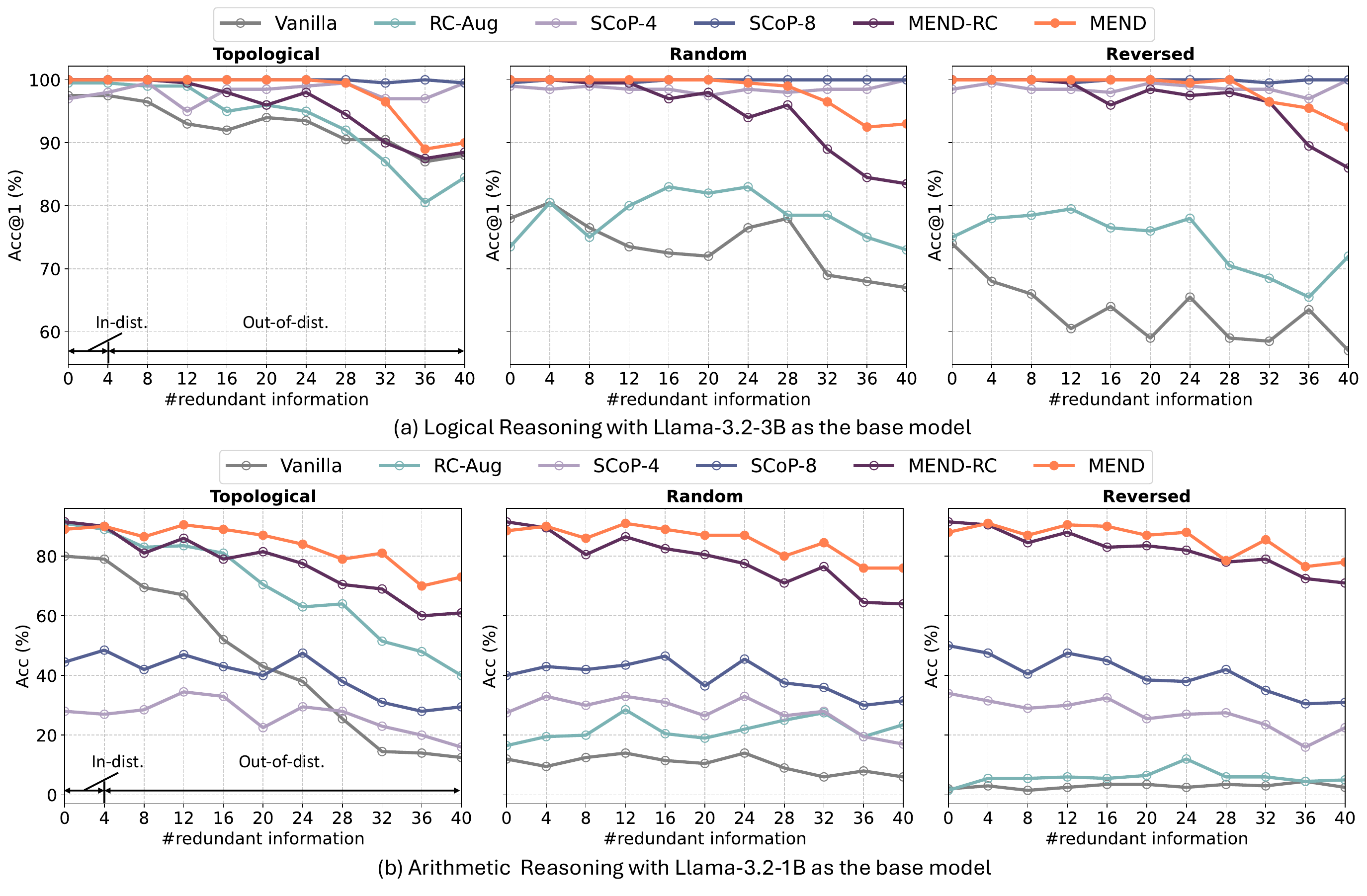}
    \caption{Evaluations with respect to different query variations. Each figure refers to one permutation order type, the x-axis represents the number of redundancies of the test set, and the y-axis represents the accuracy of final answers. For each dataset, we report the accuracy value over a dataset with a size of $200$.}
    \label{fig: Math reasoning 1B}
\end{figure*}

\subsection{Supplementary results about redundancy addition experiments}
\label{subsection:supplementary-exp-redundancy}
The results for logical reasoning with the base model {Llama-3.2-3B} and arithmetic reasoning with the base model {Llama-3.2-1B} are presented in Figure \ref{fig: Math reasoning 1B}. The findings are mostly consistent with those presented in Section~\ref{subsection:exp reasoning-equivalence enhancement}, showing that \method\ achieves the best performance. 
One exception is that the inference-time scaling baseline, \texttt{SCoP}, performs better in the logical reasoning tasks. This may be because the {Llama-3.2-3B} model is strong enough for this relatively easy task, allowing the inference-time scaling method to achieve good results through majority voting.

\subsection{Supplementary experiments about model probing}
\label{subsection: appendix-model-probing}
\begin{figure}[t]
    \centering
    \includegraphics[width=1\linewidth]{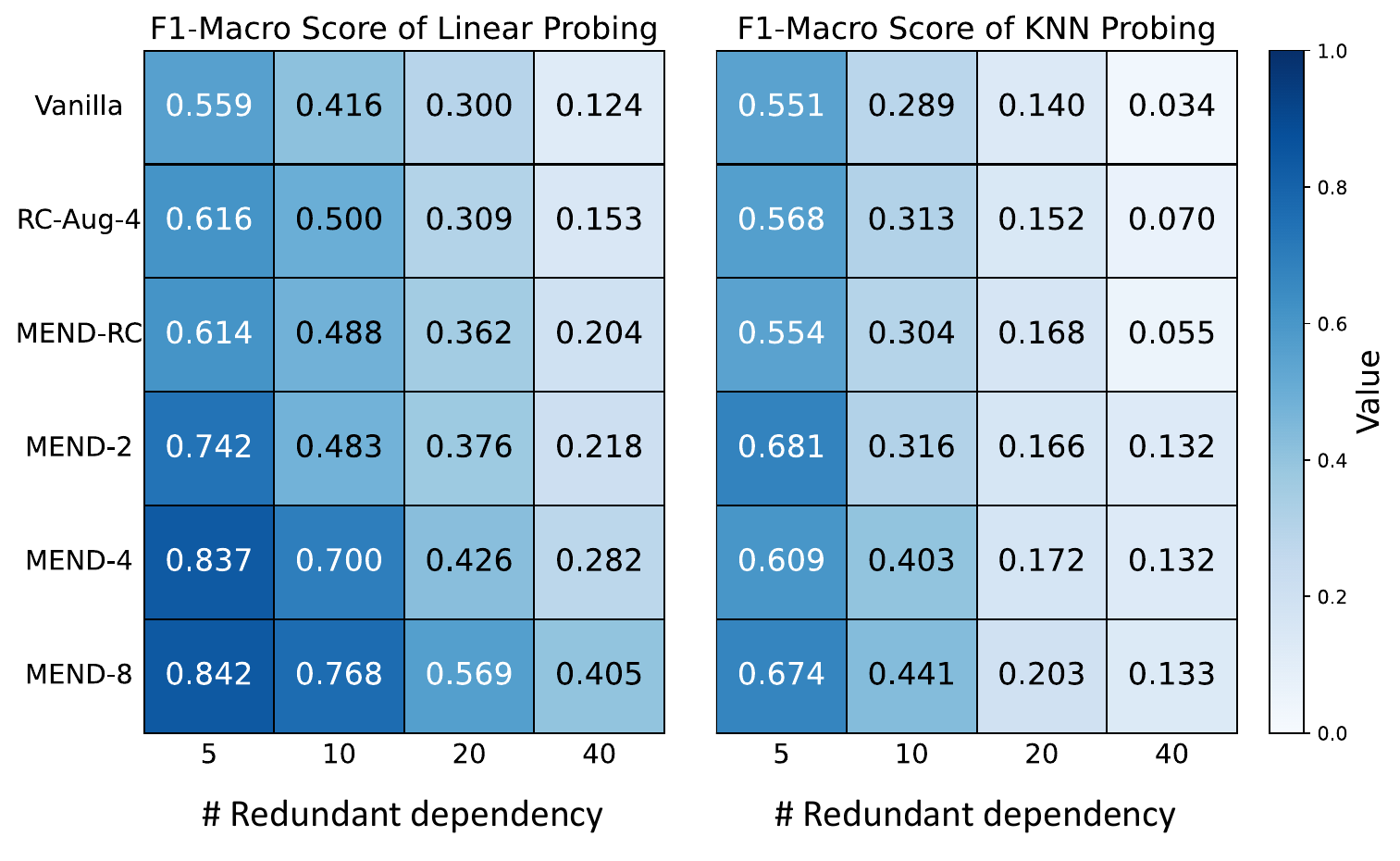}
    \caption{Comparison between linear probing and KNN-based probing used in~\citep{hou2023towards}. Left: linear probing; Right: KNN-based probing.}
    \label{fig: exp-probing-comparison}
\end{figure}

In Section~\ref{subsection: probing}, we discuss that the KNN classification method used in~\cite{hou2023towards} makes ideal assumptions and faces information aggregation issues. Here, we provide a comparison between KNN classification and the linear probing technique used in this paper. The comparison is presented in Figure~\ref{fig: exp-probing-comparison}, where we observe that although KNN probing can reveal differences between tasks with varying levels of redundancy, the F1-Macro score remains relatively low, reducing the confidence of the claims. Additionally, for harder tasks with more redundant information, the performance gap between different methods becomes minor. This is because information tends to aggregate on the beginning tokens, making it difficult to assign appropriate weights to all attention entries for information retrieval probing. Based on this experiment, we reconfirm that our linear probing technique is more suitable for revealing the model's capability.

\clearpage
\newpage

\subsection{Supplementary evaluation of closed-source models}
\label{subsection: close-source model evaluation}
The Full QA pairs presented in Figure~\ref{fig: motivation} are provided the following code blocks.




\textbf{Incorrect answer of DeepSeek-V3 to Q1.}
\begin{tcolorbox}[
   enhanced,             
   breakable,            
   colback=yellow!2!white,
   colframe=black,
   boxrule=0.5pt
]
\small
\begin{verbatim}
Let's break down the problem step by step to 
determine how many pieces of popcorn Gary 
has to eat.
1. **First 30 Seconds:**
   - **Popped Kernels:** 20
2. **Next 30 Seconds (30 to 60 seconds):**
   - The number of kernels that pop 
   is **three times** the amount that 
   popped in the first 30 seconds.
   - **Calculation:** \( 3 \times 20 = 60 \)
   - **Total Popped So Far:
   ** \( 20 + 60 = 80 \)
3. **After Removing from Heat (Residual Heat):
**
   - A **quarter** of the kernels that popped 
   in the last 30 seconds of cooking also pop 
   due to residual heat.
   - **Calculation:** \( \frac{1}{4} 
   \times 60 = 15 \)
   - **Total Popped So Far:** \( 80 + 15 = 95 \)
4. **Next 30 Seconds (60 to 90 seconds):**
   - The popping rate **increases by 20** 
   from the previous rate.
   - **Previous Rate:** 60 kernels per 30 
   seconds
   - **New Rate:** \( 60 + 20 = 80 \) kernels 
   per 30 seconds
   - **Total Popped So Far:** \( 95 + 80 = 175 \)
5. **Final 30 Seconds (90 to 120 seconds):**
   - The popping rate **slows down to half** 
   the rate of the past 30 seconds.
   - **Previous Rate:** 80 kernels per 
   30 seconds
   - **New Rate:** \( \frac{80}{2} = 40 \) 
   kernels per 30 seconds
   - **Total Popped So Far:** \( 175 + 40 = 215 \)
**Final Calculation:**
- **Total Popcorn Popped:** 215
However, let's verify the steps to 
ensure accuracy:
- **First 30 Seconds:** 
20 kernels
- **Next 30 Seconds:** \( 3 \times 20 = 60 \) 
kernels
- **Residual Heat:** 
\( \frac{1}{4} \times 60 = 15 \) kernels
- **Next 30 Seconds:** \( 60 + 20 = 80 \) kernels
- **Final 30 Seconds:** 
\( \frac{80}{2} = 40 \) kernels
Adding them up:
\[ 20 + 60 + 15 + 80 + 40 = 215 \]
Therefore, Gary has **215** pieces 
of popcorn to eat.
\[\boxed{215}\]
\end{verbatim}
\end{tcolorbox}

\newpage

\textbf{Incorrect answer of GPT-4 to Q2.}
\begin{tcolorbox}[colback=yellow!2!white, colframe=black, boxrule=0.5pt]
\small
\textit{"In the first 30 seconds, 20 kernels popped.
In the next 30 seconds, three times that amount popped, so 20 * 3 = <<20*3=60>>60 kernels popped.
In the next 30 seconds, four times the initial popping rate popped, so 20 * 4 = <<20*4=80>>80 kernels popped.
In the final 30 seconds, the popping slowed down to half the rate as the past 30 seconds, so 80 / 2 = <<80/2=40>>40 kernels popped.
After Garrett takes the pan off the heat, a quarter of the number of kernels that popped in the final 30 seconds of cooking also pop from the residual heat, so 40 / 4 = <<40/4=10>>10 kernels popped.
He believes that five kernels would not have popped if he had not shaken them to the center of the pan, so he has 20 + 60 + 80 + 40 + 10 + 5 = <<20+60+80+40+10+5=215>>215 pieces of popcorn to eat."}
\end{tcolorbox}

\clearpage
\section{Implementation Details}
In this section, we provide additional implementation details about the experiments.

\subsection{Simplified Attention}
\label{subsection: appendix-simplified_attention}
The Simplified Attention~(\ref{equ:simplified_attention}) was proposed in \texttt{MechanisticProbe}~\citep{hou2023towards}. To make this paper self-contained, we introduce some necessary details here.

Given a causal language model (LM) with $L$ layers and $H$ attention heads, the attention matrix \( A \) is represented as \( A = \{A(l, h) \mid 1 \leq l \leq L, 1 \leq h \leq H\} \), where \( A(l, h) \in \mathbb{R}^{|T| \times |T|} \). To reduce the size and complexity of \( A \) while retaining relevant information for reasoning analysis, we simplify it into \( A_{\text{simp}} \) using the following steps:

\textbf{Focus on the Last Token:} For causal LMs, attention values directed at the last input token are retained, reducing the size of \( A \) to \( A_{\text{simp}} \in \mathbb{R}^{L \times H \times |T|} \). This reduction focuses on information most relevant to the final prediction.

\textbf{Attention Head Pooling:} We apply mean pooling across all attention heads to reduce dimensionality further, resulting in \( A_{\text{simp}} \in \mathbb{R}^{L \times |T|} \).

\textbf{Hypernode Simplification (Cross-Statement Attention):} For tasks involving multi-token statements, each statement is treated as a hypernode. We apply mean pooling across tokens within each statement and max pooling across question tokens, yielding \( A_{\text{cross\_simp}} \in \mathbb{R}^{L \times (|S| + 1)} \), where \( |S| \) represents the number of statements in the query.

These simplification steps ensure that \( A_{\text{simp}} \) preserves the key information needed for probing reasoning behavior while significantly reducing computational overhead and noise. Full details are in their paper~\citep{hou2023towards}.


\subsection{Experiment Setup Details}
\label{subsection: appendix-experiment-setup}
The experiments are mainly conducted on the~\texttt{PromptBench} benchmark~\citep{zhu2023dyval}. To make the paper self-contained, we explain the data generation process for queries and responses in the experiments. Full details are provided in the documentation of \texttt{PromptBench}.


\subsubsection{Query data generation}
\begin{figure}[h]
    \centering
    \includegraphics[width=1\linewidth]{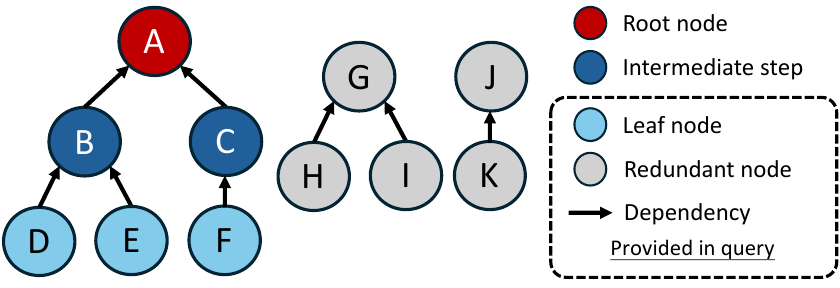}
    \caption{Illustration of the DAG structure: An example DAG with depth=$3$ and the number of redundant dependency to be $2$.}
    \label{fig: DAG illustration}
\end{figure}

The \texttt{promptbench} generates the data in two stages: (1) DAG construction; and (2) Natural Language Description of the DAG. An illustration of the DAG is shown in FIgure~\ref{fig: DAG illustration}.

\textbf{DAG construction:} We generate the DAG with a specified depth and number of redundancy. Based on these, we generate the DAG from top to bottom: Firstly generate the root node, and then sample the dependency between itself and its parent node(s). If the dependency contains a $2$-variable operator (e.g. \{$+$, $-$, $\land$ (\texttt{AND})\}), then it has two parent nodes to generate, otherwise, if it contains a $1$-variable operator (e.g. \{$\square^2$, $\neg$ (\texttt{NOT})\}), it only has one parent node. After the generation of the root node, we then go to the parent generation of its parent node(s), and do this recursively until reach the expected depth. Along with the DAG generation, we also obtain the names of all the generated nodes from a random string generator.
After obtaining the DAG, we sample the values for all the leaf nodes from the pre-defined set, and calculate the value of their child nodes from bottom to top.

\textbf{Natural Language Description}
After obtaining the DAG and all the node information, we use natural language to describe the question with pre-defined templates. If it is the leaf node, we describe its name and value as: 
\begin{tcolorbox}[colback=blue!2!white, colframe=blue!80!black, boxrule=0.5pt]
\small
\textit{"The value of} \texttt{node.name} \textit{is} \texttt{node.value}\textit{."}
\end{tcolorbox}

If it is an intermediate node or root node, we describe the dependency with its parent nodes as:

\begin{tcolorbox}[colback=blue!2!white, colframe=blue!80!black, boxrule=0.5pt]
\small
\textit{"}\texttt{node.name} \textit{gets its value by}  \texttt{template}-\texttt{func}(\texttt{node.operator}, \texttt{node.parent})\textit{."}
\end{tcolorbox}
where \texttt{template}-\texttt{func} is a template function to generate strings based on the operator. For example, if the operator is ``$-$'', then the generated string from \texttt{template}-\texttt{func} is:

\begin{tcolorbox}[colback=blue!2!white, colframe=blue!80!black, boxrule=0.5pt]
\small
\textit{"subtracting the value of} \texttt{node.parent1.name} from the value of \texttt{node.parent2.name}\textit{."}
\end{tcolorbox}

If all the information is described in the \texttt{topological} sorting order, then it forms the \texttt{topological} order dataset. If the sentences are shuffled, it forms the \texttt{random} dataset. If the sentences are reversed permutated, then it forms the \texttt{reversed} dataset. From the templates above, we can observe that permutating the sentence order in the question prompts does not change the overall semantic meaning of this query.

For the redundancy generation, we describe the redundant nodes and their corresponding dependency based on the process and templates describe above. Since the redundant information is not related to the root node, and their names are not overlapped, describing the redundant information also does not change the semantic meaning of the useful information.

At the end of the problem description, we add one sentence to conclude the question:
\begin{tcolorbox}[colback=blue!2!white, colframe=blue!80!black, boxrule=0.5pt]
\small
\textit{"What is the value of} \texttt{root.name}\textit{?"}
\end{tcolorbox}

\label{subsection: example-qa-pairs}
We provide example QA pairs for the arithmetic reasoning and logic reasoning tasks and their corresponding DAG structure in this part.
\begin{figure}[t]
    \centering
    \includegraphics[width=1\linewidth]{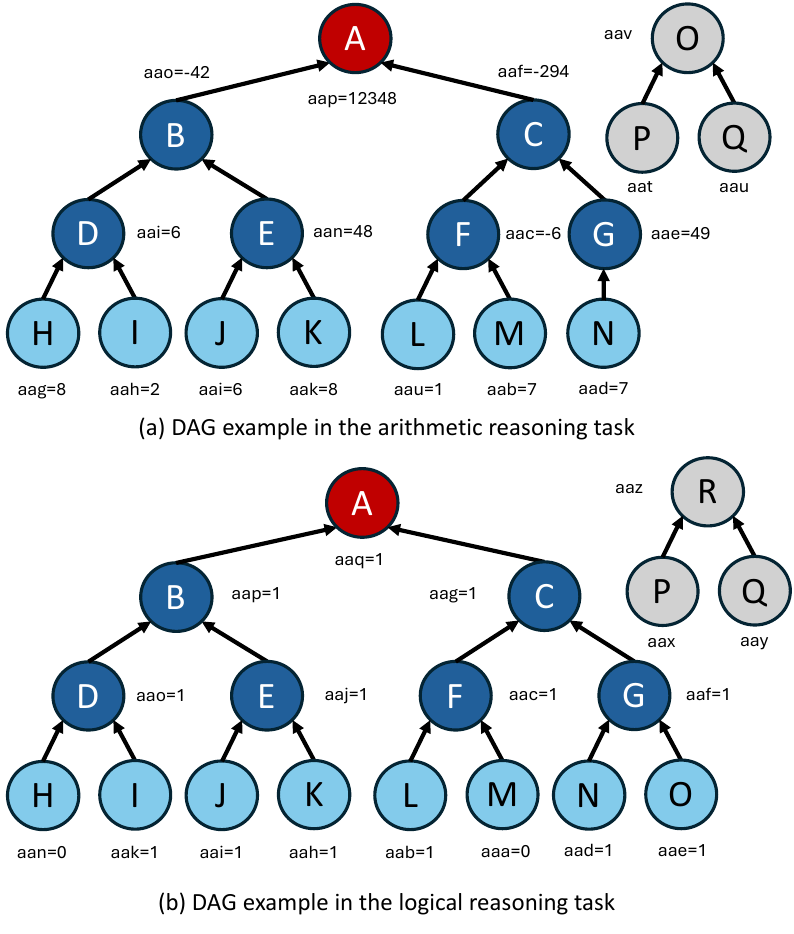}
    \caption{DAG Example in the arithmetic reasoning task and the logical reasoning task.}
    \label{fig: DAG_math_example}
\end{figure}

\subsubsection{Response generation}
In addition to the query mentioned above, SFT data also contains the description of the reasoning chain. As mentioned in section~\ref{section: preliminary}, the ground-truth reasoning chain is a topological sorting of this DAG. Similar to the process in query generation, the reasoning chain generation also utilizes some templates. The difference is that, it contains the answer after every intermediate and final reasoning step. For example:

\begin{tcolorbox}[colback=blue!2!white, colframe=blue!80!black, boxrule=0.5pt]
\small
\textit{"}\texttt{node.name} \textit{gets its value by}  \texttt{template}-\texttt{func}(\texttt{node.operator}, \texttt{node.parent})\textit{, so the value of} \texttt{node.name} \textit{is} \texttt{node.value} \textit{."}
\end{tcolorbox}

\subsection{Example QA Pairs}
\label{subsection:appendix-example-QA-pairs}

For the DAG example shown in Figure~\ref{fig: DAG_math_example}, the corresponding QA pairs are shown in the following text blocks:

\textbf{Arithmetic reasoning:}

\begin{tcolorbox}[colback=yellow!2!white, colframe=black, boxrule=0.5pt]
\small
\textit{Question: \\ "The value of aag is 8.\\ The value of aah is 2.\\ aai gets its value by subtracting the value of aah from the value of aag.\\ aav gets its value by multiplying together the value of aat and aau.\\ The value of aaj is 6.\\ The value of aat is 9.\\ The value of aak is 8.\\ aan gets its value by multiplying together the value of aaj and aak.\\ aao gets its value by subtracting the value of aan from the value of aai.\\ The value of aau is 1.\\ The value of aab is 7.\\ The value of aad is 7.\\ The value of aaa is 1.\\ aac gets its value by subtracting the value of aaa from the value of aab.\\ aae gets its value by squaring the value that aad has.\\ aaf gets its value by multiplying together the value of aac and aae.\\ aap gets its value by multiplying together the value of aaf and aao.\\ What is the value of aap?"}
\end{tcolorbox}

\begin{tcolorbox}[colback=yellow!2!white, colframe=black, boxrule=0.5pt]
\small
\textit{Response: \\ "aag is 8.0\\ aah is 2.0\\ aai = aag - aah = 8.0 - 2.0 = 6.0\\ aaj is 6.0\\ aak is 8.0\\ aan = aaj * aak = 6.0 * 8.0 = 48.0\\ aao = aai - aan = 6.0 - 48.0 = -42.0\\ aab is 7.0\\ aad is 7.0\\ aaa is 1.0\\ aac = aab - aaa = 7.0 - 1.0 = -6.0\\ aae = $aad^2$ = $(7.0)^2$ = 49.0\\ aaf = aac $*$ aae = -6.0 $*$ 49.0 = -294.0\\ aap = aao $*$ aaf = -42.0 $*$ -294.0 = 12348.0\\ Thus, the answer is 12348.0"}
\end{tcolorbox}

\textbf{Logical reasoning:}

\begin{tcolorbox}[colback=yellow!2!white, colframe=black, boxrule=0.5pt]
\small
\textit{Question: \\ "The value 1 means True, and the value 0 means False.\\ aak is 1.\\ aai is 1.\\ aah is 1.\\ The value of aaj equals to (aai AND aah).\\ aan is 0.\\ aax is 1.\\ The value of aao equals to (aak OR aan).\\ The value of aap equals to (aaj OR aao).\\ aab is 1.\\ aaa is 0.\\ The value of aac equals to (aab OR aaa).\\ aad is 1.\\ aae is 1.\\ The value of aaz equals to (aax AND aay).\\ The value of aaf equals to (aad AND aae).\\ The value of aag equals to (aaf AND aac).\\ The value of aaq equals to (aap AND aag).\\ aay is 1.\\ What is the value of aaq?"}
\end{tcolorbox}

\begin{tcolorbox}[colback=yellow!2!white, colframe=black, boxrule=0.5pt]
\small
\textit{Response: \\ "aak is 1.\\ aan is 0.\\ aao = (aak OR aan) = (1 OR 0) = 1.\\ aah is 1.\\ aai is 1.\\ aaj = (aah AND aai) = (1 AND 1) = 1.\\ aap = (aao OR aaj) = (1 OR 1) = 1.\\ aab is 1.\\ aaa is 0.\\ aac = (aab OR aaa) = (1 OR 0) = 1.\\ aad is 1.\\ aae is 1.\\ aaf = (aad AND aae) = (1 AND 1) = 1.\\ aag = (aaf AND aac) = (1 AND 1) = 1.\\ aaq = (aag AND aap) = (1 AND 1) = 1.\\ Thus, the answer is 1"}
\end{tcolorbox}


\subsection{Baseline Implementation Details}
\begin{itemize}
    \item \texttt{SCoP}-$k$. First, it paraphrases the question prompts \( k \) times, then performs reasoning based on these paraphrases, and obtains the final answer through majority voting. Unlike the closed-source LLMs used in~\cite{zhou2024paraphrase}, our LLMs after SFT lose some of their paraphrasing capability. Additionally, in our experiment, sentence order permutation does not alter the overall semantic meaning, which is a good way for paraphrase. Therefore, we permute the sentences to achieve paraphrasing, resulting in better final performance.

    \item \texttt{RC-Aug} utilizes the reasoning chain (RC) augmented dataset for SFT, as used in previous work~\citep{yu2023metamath}. Specifically, we augment the answers with different topological orderings while keeping the query unchanged. For a DAG with more than two levels, the topological ordering is not unique. For example, in the DAG shown in Figure~\ref{fig: DAG illustration}, the topological ordering can start by solving either node B first or node C first. When solving node B, it can begin with either node D or node E. Consequently, a fixed DAG can have multiple valid topological orderings, resulting in different reasoning chains that lead to the same correct answer. In \texttt{RC-Aug}, we incorporate these additional reasoning chains from different topological orderings to augment the SFT dataset.

    \item \texttt{MEND-RC} is an ablation variant that applies \method\ to transform the original question queries without introducing new queries into the dataset, while augmenting the dataset with additional reasoning paths.
\end{itemize}

All the methods include the baselines and \method\ utilize full-parameter fine-tuning. We report the accuracy as Pass@1 if not indicated otherwise.

\subsection{Computing Resources.}
The experiments are run on a server with 2$\times$AMD EPYC 7542 32-Core Processor CPU, 2$\times$NVIDIA RTX A6000 graphics, and $252$ GB memory. For the arithmetic reasoning tasks with 3B model, it takes about $20$ GPU hours for SFT and evaluation. For Logical reasoning tasks with 1B model, it takes about $4$ GPU hours for SFT and evaluation.


\end{document}